\documentclass[letterpaper, journal]{IEEEtran}

\makeatletter
\let\NAT@parse\undefined
\makeatother

\newif\ifanonymize
\anonymizefalse 

\usepackage{amsmath, amsfonts, amssymb}

\usepackage{array}          
\usepackage{multirow}
\usepackage{threeparttable}
\usepackage{booktabs}

\usepackage{graphicx}
\usepackage{caption}
\usepackage{enumitem}
\usepackage[table, dvipsnames]{xcolor}

\usepackage[noadjust]{cite}
\usepackage{url}
\usepackage{hyperref}
\usepackage{cleveref}       

\usepackage{xspace}
\usepackage{pifont}

\providecommand{\citet}[1]{\cite{#1}}
\providecommand{\citeauthor}[1]{\cite{#1}}

\hypersetup{
  colorlinks = true,
  citecolor  = Green,
  linkcolor  = NavyBlue,
  filecolor  = OrangeRed,
  urlcolor   = OrangeRed,
  pdfpagemode = FullScreen,
}

\newcommand{\papertitle}{\framework: Towards Standardizing Robustness Evaluation in Trajectory Prediction Under Distribution Shifts}

\newcommand{\idest}{i.e.,\xspace}
\newcommand{\exempli}{e.g.,\xspace}
\newcommand{\etal}{et~al.\@\xspace}

\newcommand{\paragraphbf}[1]{\vspace{0.15cm}\noindent\textbf{#1.}\xspace}

\colorlet{SeenColor}{ForestGreen}
\colorlet{UnseenColor}{Red}
\colorlet{VehicleColor}{Black!85}
\colorlet{PedestrianColor}{Magenta!85}
\colorlet{CyclistColor}{ForestGreen!85}
\colorlet{EgoAgentColor}{RoyalBlue!85}
\colorlet{CausalColor}{Orange!85}
\colorlet{BackgroundColor}{Orange!85}
\colorlet{WomdColor}{Black}
\colorlet{MyFrameworkColor}{Black!70}
\colorlet{OtherFrameworkColor}{Black}
\definecolor{NameColor}{RGB}{0,0,0}   

\newcommand{\mbf}[1]{{\mathbf #1}}
\newcommand{\state}{{\mathbf x}}
\newcommand{\context}{{\mathbf C}}
\newcommand{\meta}{{\mathbf M}}
\newcommand{\setS}{{\mathcal{S}}}
\newcommand{\setB}{{\mathcal{B}}}
\newcommand{\setM}{{\mathcal{M}}}

\newcommand{\seenS}{{\setS_{\text{SEEN}}}}
\newcommand{\unseenS}{{\setS_{\text{UNSEEN}}}}

\newcommand{\seen}{{\textcolor{SeenColor}{\textsc{seen}}}}
\newcommand{\unseen}{{\textcolor{UnseenColor}{\textsc{unseen}}}}

\newcommand{\egoagent}{{\textcolor{EgoAgentColor}{\text{ego-agent}}}}

\newcommand{\myframeworkname}[1]{{\textcolor{MyFrameworkColor}{\textsc{#1}}}\xspace}
\newcommand{\otherframeworkname}[1]{{\textcolor{OtherFrameworkColor}{\textsc{#1}}}\xspace}
\newcommand{\modelname}[1]{{\textcolor{OtherFrameworkColor}{\textsc{#1}}}\xspace}
\newcommand{\benchmarkname}[1]{{\textcolor{MyFrameworkColor}{\textsc{#1}}}\xspace}

\newcommand{\framework}{\myframeworkname{ControlledShifts}}
\newcommand{\carnovel}{\otherframeworkname{CarNovel}}
\newcommand{\longcomp}{\otherframeworkname{LongComp}}
\newcommand{\frenetp}{\otherframeworkname{Frenet+}}

\newcommand{\safeshift}{\otherframeworkname{SafeShift}}

\newcommand{\netlsd}{\textsc{NetLSD}\xspace}
\newcommand{\networkx}{\textsc{NetworkX}\xspace}

\newcommand{\womd}{{\textcolor{WomdColor}{\textsc{WOMD}}}}

\newcommand{\naive}{\modelname{Naive}}
\newcommand{\autobot}{\modelname{AutoBot}}
\newcommand{\scenetransformer}{\modelname{SceneTransformer}}
\newcommand{\wayformer}{\modelname{Wayformer}}
\newcommand{\mtr}{\modelname{MTR}}

\newcommand{\backgroundagents}{\benchmarkname{BackgroundAgents}}
\newcommand{\egosafeshift}{\benchmarkname{EgoSafeShift}}
\newcommand{\environments}{\benchmarkname{Environments}}
\newcommand{\uniform}{\benchmarkname{Uniform}}
\newcommand{\backgroundagentsabv}{\benchmarkname{bga}}
\newcommand{\egosafeshiftabv}{\benchmarkname{ess}}
\newcommand{\environmentsabv}{\benchmarkname{env}}
\newcommand{\uniformabv}{\benchmarkname{uni}}
\newcommand{\causalagents}{\otherframeworkname{CausalAgents}}

\def\HiLiYellow{\leavevmode\rlap{\hbox to \hsize{\color{yellow!20}%
  \leaders\hrule height .8\baselineskip depth .4ex\hfill}}}

\newif\ifshowstatshorizontal
\showstatshorizontaltrue

\graphicspath{{figures/}}
\newcommand{\includegraphicsdpi}[3]{
    \pdfimageresolution=#1  
    \includegraphics[#2]{#3}
    \pdfimageresolution=20  
}

\title{\papertitle}

\ifanonymize
\author{Authors Omitted for Review}
\else
\author{
Ingrid Navarro$^{1,2}$, 
Pablo Ortega-Kral$^{1,2}$, 
Yutong Duan$^{3}$,
Jonathan Francis$^{1,4,\dag}$, 
and Jean Oh$^{1,\dag}$
\thanks{$^{1}$Robotics Institute, School of Computer Science, Carnegie Mellon University. {\tt\scriptsize \{ingridn, portegak, jeanoh\}@cs.cmu.edu}}
\thanks{$^{2}$Work done as part of an internship at Lavoro AI.}
\thanks{$^{3}$Stack AV}
\thanks{$^{4}$Bosch Center for Artificial Intelligence: {\tt\scriptsize jon.francis@us.bosch.com}}
\thanks{$^{\dag}$Equal Advising.}
}
\fi


\let\oldtwocolumn\twocolumn
\renewcommand\twocolumn[1][]{%
    \oldtwocolumn[{#1}{
    \begin{center}
           \vspace{-1cm}
           \includegraphicsdpi{200}{width=0.98\textwidth}{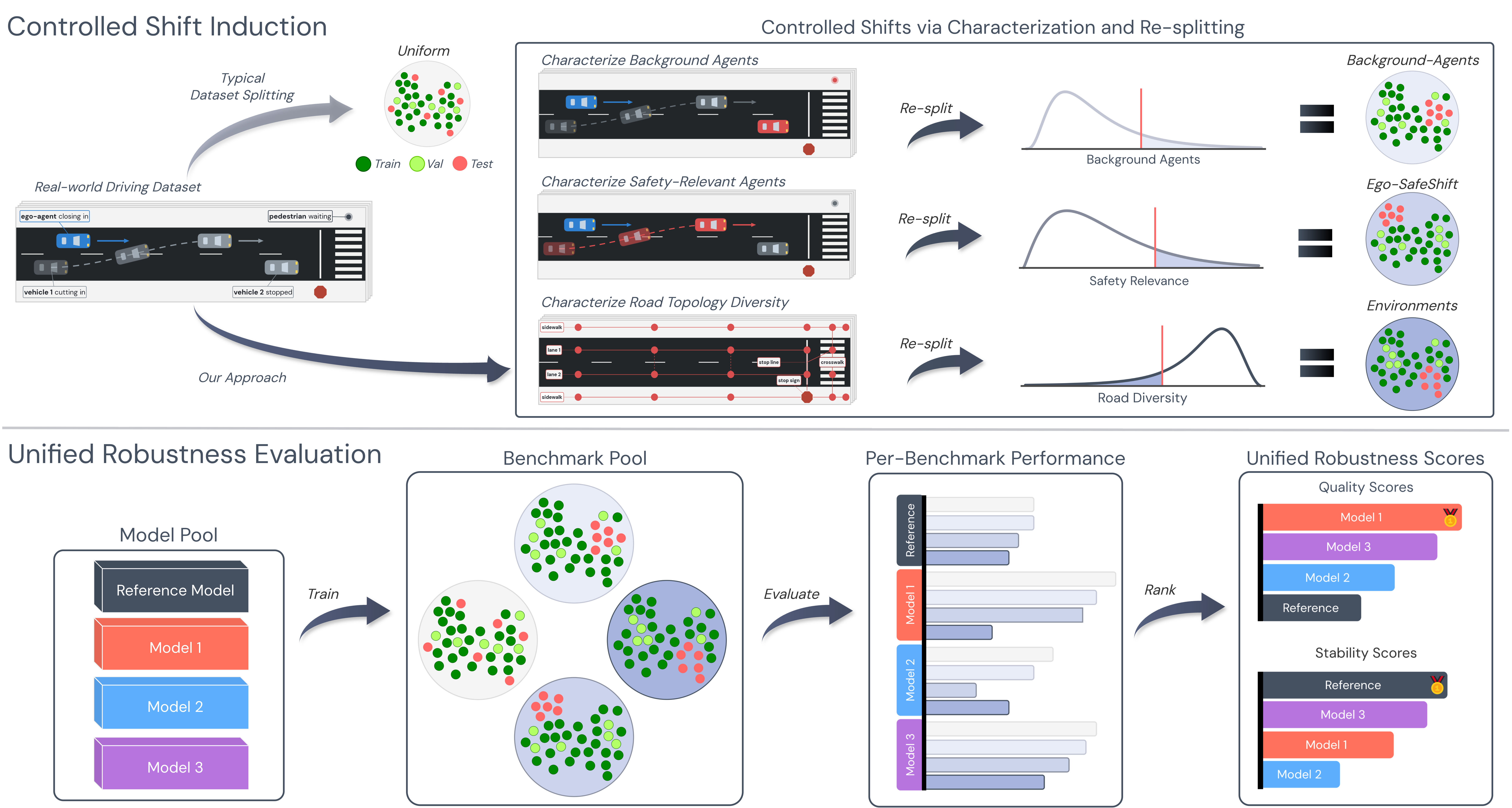}
           \captionof{figure}{Overview of \framework. \textbf{Controlled Shift Induction} (top): a characterization function $c$ maps each scenario to a desired representation, and a splitting function $f$ withholds the tail of the resulting distribution, giving \seen (train/val) and \unseen (test) partitions, instead of a uniform random split. 
           Scenarios such as the toy cut-in example shown here are characterized three ways: by background-agent count, safety relevance, and road diversity, yielding \backgroundagents, \egosafeshift, and \environments. The same corpus, but three different distributions and held-out tails. \textbf{Unified Robustness Evaluation} (bottom): trajectory predictors are then trained and evaluated on this benchmark pool, and their results are unified into \emph{quality} and \emph{stability} robustness scores.}
           \label{fig:overview}
        \end{center}
    }]
}

\begin{document}

\maketitle
\bstctlcite{IEEEexample:BSTcontrol}
%

\renewcommand{\thefootnote}{\arabic{footnote}}

\begin{abstract}
Trajectory prediction is central to safety in autonomous driving, yet learning-based predictors tend to degrade sharply when encountering scenarios poorly represented by their training data. Many methods attempt to mitigate distribution shift degradation through data-centric or test-time adaptation approaches; however, they are typically validated along fragmented axes of generalization, leaving the field without a standardized way to compare robustness across shifts a model may encounter. 

To address this, we introduce \framework, a framework and benchmark suite that systematically re-splits existing trajectory datasets into in-distribution (\seen) and out-of-distribution (\unseen) partitions, via a shared characterization-and-splitting formulation, in which a characterization function fixes the axis of variation a benchmark probes and a splitting function fixes how the tail of that axis is withheld. The suite comprises three benchmarks targeting key topological and behavioral distribution shifts. Furthermore, to aggregate multi-dimensional performance metrics across these benchmarks, we propose a unified robustness score that evaluates models along two complementary dimensions: prediction quality (relative performance gain) and prediction stability (performance preservation under shift). We showcase \framework~by benchmarking prominent transformer-based architectures, exposing critical differences in how models of varying capacities handle latent relevance and environmental structure.

\ifanonymize
\else
Our code and benchmarks are publicly available at \href{https://navars.xyz/controlledshifts/}{navars.xyz/controlledshifts/}.
\fi
\end{abstract}


\section{Introduction} 
\label{sec:introduction}

Data-driven trajectory prediction has become a core task for safety in autonomous driving (AD) and, more broadly, multi-agent settings. Recent methods~\cite{shi2022motion, nayakanti2023wayformer, navarro2024amelia, ngiam2021scene} have excelled at the task by leveraging large training datasets~\cite{ettinger2021large, navarro2024amelia, caesar2020nuscenes, wilson2023argoverse, huang2026nureasoning} spanning diverse real-world scenarios and interactions. At deployment time, however, autonomous systems inevitably face novel situations their training data covers poorly: unseen road topologies, unrelated background actor behavior, unexpected maneuvers, or erratic driving. Safety demands that a model respond sensibly to these cases, yet learning-based methods generally degrade under such distribution shifts~\cite{stoler2024safeshift, stoler2025longcomp, li2024adaptive}.

Efforts to improve robustness to distribution shifts fall into two categories: the first targets the data, \exempli~via scenario mining~\cite{stoler2024safeshift, stoler2025longcomp, navarro2024amelia} or via coverage widening through scenario generation~\cite{stoler2025seal, stoler2025rcg, huang2025cadre}; the second acts at test time, handling degraded predictions through uncertainty estimation~\cite{ivanovic2022propagating}, model ensembles~\cite{li2024adaptive}, or gradient-based shift detection~\cite{de2026forecasting}. Despite this activity, measuring how robust a model actually is remains an \textit{ad hoc} exercise. Each method commits to its own axis of generalization~\cite{stoler2024safeshift, stoler2025longcomp, ye2023frenet}, making comparison difficult. With no shared basis for evaluation, robustness cannot be contextualized across methods and shift conditions. In fields such as robotics, benchmarks that aggregate distinct sources of variation have proven valuable for quantifying robustness and guiding model development~\cite{zhou2025libero, koh2021wilds}; trajectory prediction in autonomous driving lacks definitive efforts towards such standardization.

To address this gap, we present \framework~(\Cref{fig:overview}), a framework and benchmark suite for standardized evaluation of the robustness of trajectory prediction models under distribution shift. Rather than collecting new data, \framework~re-splits an existing dataset into \seen~and \unseen~partitions through a shared \textit{characterization-and-splitting} formulation, instantiated across distinct axes of variation. This yields a consistent protocol under which different approaches can be measured and compared directly.

To summarize, our contributions are as follows:
\begin{itemize}[leftmargin=1.2em, itemsep=1pt, topsep=2pt, parsep=0pt]
    \item We \textbf{formalize and unify distribution-shift creation}, combining a characterization function $c$ and a splitting function $f$ (\Cref{ssec:inducing_shifts}), providing a single, standardized formulation for inducing controlled shifts from existing data across heterogeneous axes of variation.
    \item We propose a \textbf{unified robustness score} for standardized comparison across models (\Cref{ssec:unified_robustness}), which aggregates benchmarks' \seen/\unseen~performance, condensing nuanced, multi-dimensional measurements into robustness values along two dimensions, \textit{quality} and \textit{stability}.
    \item We instantiate our formulation as \textbf{three complementary benchmarks} that target behavioral and topological shifts relevant to the \egoagent~(\Cref{sec:benchmark_suite}): \backgroundagents, \egosafeshift, and \environments, and showcase our \textbf{robustness scheme} on key trajectory prediction models (\Cref{sec:evaluation}).
\end{itemize}

\section{Related Work}
\label{sec:related_works}

\subsection{Distribution Shift Benchmarks in Trajectory Prediction}
\label{ssec:distribution_shift_benchmarks}

Various works study performance under distribution shift conditions. One direction targets test-time detection and adaptation: \carnovel~\cite{filos2020can} proposes an uncertainty-aware planner that detects and recovers from out-of-distribution scenes, with a benchmark for topological shifts, while \causalagents~\cite{sun2024causalagents} assesses sensitivity to non-causal perturbations.

A complementary direction induces shifts by re-splitting existing datasets with respect to a chosen basis.
\frenetp~\cite{ye2023frenet} clusters scenarios by environmental features such as lane deflection and geometry, holds out a subset of clusters for testing, and applies Frenet normalization as remediation. Shifting the basis from environmental features to safety-relevance, \safeshift~\cite{stoler2024safeshift} mines subtly risky scenarios via a safety-score function, and \longcomp~\cite{stoler2025longcomp} factorizes scenarios into ego and social contexts, holding out novel combinations to build compositional, zero-shot settings.

Each of these commits to a single, fixed basis for partitioning: environmental, safety-critical, or compositional. To our knowledge, no prior work decouples the \emph{characterization} of scenarios from the \emph{re-splitting} strategy. We treat shift creation as a general formulation that can be instantiated along arbitrary axes of disentanglement. To ensure the resulting shifts are meaningful rather than arbitrary, we instantiate our method on axes that prior research has deemed relevant, namely topological and behavioral, and validate via downstream evaluation in \Cref{sec:evaluation}  and distribution analyses in Appendix \ref{sec:benchmark_analysis}.

\subsection{Measuring Robustness in Trajectory Prediction}
\label{ssec:measuring_robustness}

One line of robustness evaluation measures metric degradation under adversarial attacks~\cite{weng2023joint, saadatnejad2022socially, zhang2022adversarial}, showing that models often lack basic safety and social awareness. Because these methods simulate adversarial agent behavior, however, the resulting scenarios inherit a simulation-to-real gap: degradation may reflect implausible simulated dynamics rather than genuine model weakness~\cite{hanselmann2022king, xu2023bits, francis2022core, huang2023went}.

Other works assess the degradation of standard metrics (\exempli displacement error, miss rate, collision rate) on held-out sets~\cite{stoler2024safeshift, ye2023frenet, zhang2022adversarial, cao2022advdo}. These sets are defined by desired scenario characteristics, such as road shape~\cite{filos2020can, ye2023frenet}, side-of-driving~\cite{itkina2023interpretable}, motion style~\cite{shi2022motion, itkina2023interpretable}, and safety factors~\cite{stoler2024safeshift, stoler2025longcomp}. Many also introduce degradation mitigation strategies, \exempli Frenet coordinates~\cite{ye2023frenet, werling2010optimal}, few-shot adaptation~\cite{filos2020can}, motion style transfer~\cite{kothari2023motion}, or safety-awareness~\cite{stoler2024safeshift}.
 
Yet these results are reported per shift, leaving the field without a metric that captures how robust a model is across the shifts it may encounter. We address this with a robustness scoring scheme along two dimensions, \textit{quality} and \textit{stability}, defined in \Cref{ssec:unified_robustness}.

\section{Controlled Shifts and Unified Robustness Formulation}
\label{sec:formulation}

We examine the robustness of trajectory forecasting models to targeted distribution shifts in urban driving scenarios. Our formulation below builds upon that of Stoler \etal~\cite{stoler2024safeshift}.

\subsection{Preliminaries}
\label{ssec:preliminaries}

\paragraphbf{Trajectory Prediction} Let $\setS$ denote the set of scenarios comprising a motion prediction dataset. A scenario $s = (\mathbf{X}, \context, \meta) \in \setS$ consists of the trajectories of all observed agents $\mathbf{X} = \{\mathbf{x}^{(i)}\}_{i=1}^{N}$, map information $\context$, and meta information $\meta$. 
Each trajectory $\mathbf{x}^{(i)} = (\state^{(i)}_1, \ldots, \state^{(i)}_{T_{\text{tot}}})$ is a sequence of agent states, where $\state^{(i)}_t$ denotes the state of agent $i$ at timestep $t$. 
The map $\context$ contains road information, \exempli lane locations and connectivity, while $\meta$ provides additional task specifications, such as the set of agents to be predicted (\idest~\textit{target} agents).
Each trajectory is partitioned at the final observed timestep $T_{\text{obs}}$ into a history and a future segment:
\[
\mathbf{x}^{(i)}_{\text{hist}} = (\state^{(i)}_1, \ldots, \state^{(i)}_{T_{\text{obs}}}), \quad
\mathbf{x}^{(i)}_{\text{fut}} = (\state^{(i)}_{T_{\text{obs}}+1}, \ldots, \state^{(i)}_{T_{\text{tot}}}).
\]

Writing $s_{\text{hist}} = (\mathbf{X}_{\text{hist}}, \context, \meta)$ for the observable portion of the scenario, the trajectory prediction goal is to estimate the future trajectories $\mathbf{x}^{(i)}_{\text{fut}}$ of the target agents given $s_{\text{hist}}$.

\subsection{Inducing Controlled Distribution Shifts}
\label{ssec:inducing_shifts}

To create a benchmark $b \in \setB$, we partition $\setS$ into disjoint subsets $\seenS$ and $\unseenS$. Let $c: \setS \rightarrow \mathcal{E}$ be a characterization function that maps each scenario $s$ to its representation $e = c(s)$. Then, a splitting function $f: \mathcal{E} \rightarrow \{\seen, \unseen\}$ assigns each scenario to a subset based on its representation, relative to the full corpus, inducing the partition $\setS = \seenS \cup \unseenS$. Models are trained and validated exclusively on $\seenS$, while $\unseenS$ is held out for testing.
 
Here $c$ determines the axis of variation a benchmark probes and $f$ how its tail is withheld. Thus, either can be replaced without disturbing the other. This formulation has three main benefits. First, collecting new data is not required since a benchmark is a re-splitting of existing data, 
so scenario format and quality are held fixed, making the induced shift the only variation. Second, benchmarks built from different characterization functions become comparable, as they are re-splits of the same underlying corpus and are scored by the same scheme (\Cref{ssec:robustness_scores}). Third, the suite is extensible: adding an axis of variation requires only a new $(c, f)$ pair.


\subsection{Unified Robustness under Distribution Shifts} 
\label{ssec:unified_robustness}

\subsubsection{Dimensions of Robustness} Given an induced shift, the goal is to minimize performance degradation when models trained and validated exclusively on $\seenS$ are evaluated on $\unseenS$. Degradation on its own, however, is insufficient: a weak predictor can degrade little simply because it had little to lose. We therefore measure robustness along two complementary dimensions, each fixed by the reference a predictor is compared against:

\paragraphbf{Quality} This dimension captures downstream performance by measuring a model against a \naive~baseline within a benchmark. Here, we instantiate a predictor given neither map nor social context (\Cref{ssec:experimental_setup}).

\paragraphbf{Stability} This dimension captures how well a model maintains its performance under shift, by measuring it against its counterpart trained on a \uniform~setting, \idest when no controlled shift is applied. 

\subsubsection{Robustness Scores} 
\label{ssec:robustness_scores}
Inspired by prior work~\cite{hyndman2006another, koh2021wilds, sagawa2020distributionally}, we define the overall robustness of a model as follows:

\paragraphbf{Split Ratio} For a metric $m$, predictor $p$, reference model $r$, and benchmark $b$, the split-level ratio compares the reference's score against the predictor's on that split:
\begin{align*}
    \rho^{(m,\,p,\,r)}_{b,\,\textsc{split}} &= \frac{m^{(r,\,b)}_{\textsc{split}}}{m^{(p,\,b)}_{\textsc{split}}}, \quad \textsc{split} \in \{\seen, \unseen\}
\end{align*}
Every metric we consider is an error measure, so a ratio above $1$ indicates that the predictor outperforms the reference. The choice of $r$ selects the dimension, \idest setting $r$ to the \naive~baseline yields \emph{quality}, while setting it to the predictor's own score on the \uniform setting yields \emph{stability}. 

\paragraphbf{Per-benchmark Score} We combine the split scores via a geometric mean and average them across metrics $\setM$ into a score per predictor, benchmark, and dimension:
\begin{align*}
    \mbf{r}^{(p,\,r)}_{b} &= \frac{1}{|\setM|}\sum_{m \in \setM} \sqrt{\rho^{(m,\,p,\,r)}_{b,\,\seen} \cdot \rho^{(m,\,p,\,r)}_{b,\,\unseen}}
\end{align*}
These are the robustness values reported in \Cref{tab:distribution_shift_results}. The geometric mean penalizes a predictor's performance imbalance. 

\paragraphbf{Unified Robustness Score} Finally, we average the per-benchmark scores over the shifted benchmarks, giving one value per predictor and dimension:
\begin{align*}
    \mbf{R}^{(p,\,r)} &= \frac{1}{|\setB_{\text{shift}}|}\sum_{b \in \setB_{\text{shift}}} \mbf{r}^{(p,\,r)}_{b}, \quad \setB_{\text{shift}} = \setB \setminus \{\text{\uniform}\}
\end{align*}
We use the unified scores to report the rankings in \Cref{fig:model_scores}.

\section{The \framework Benchmark Suite}
\label{sec:benchmark_suite}

We instantiate our formulation as three benchmarks, each targeting a different aspect in which a scenario can be \textit{interesting} to the \egoagent.
\backgroundagents (\backgroundagentsabv) probes robustness to distractor agents, \idest whether a model identifies agents that are relevant to the \egoagent's decision and ignores those that are not.
\egosafeshift (\egosafeshiftabv) probes safety relevance, \idest nuanced, latent criticality that the \egoagent~must anticipate to prevent accidents. 
\environments (\environmentsabv) probes structural diversity, \idest whether representations learned over varied road topologies transfer to held-out topologies. 

As detailed below, each benchmark instantiates its targeted characterization $c_{b}$ and splitting $f_{b}$ functions from \Cref{ssec:inducing_shifts}, differing in the scenario properties used to induce the shift. \Cref{fig:distribution_shift_benchmarks} shows representative examples for each, and in Appendix~\ref{sec:benchmark_analysis}, we provide data analyses across each benchmark to further validate the design choices.

\begin{figure}[!ht]
    \centering
    \vspace{0.2cm}
    \includegraphics[width=0.96\linewidth]{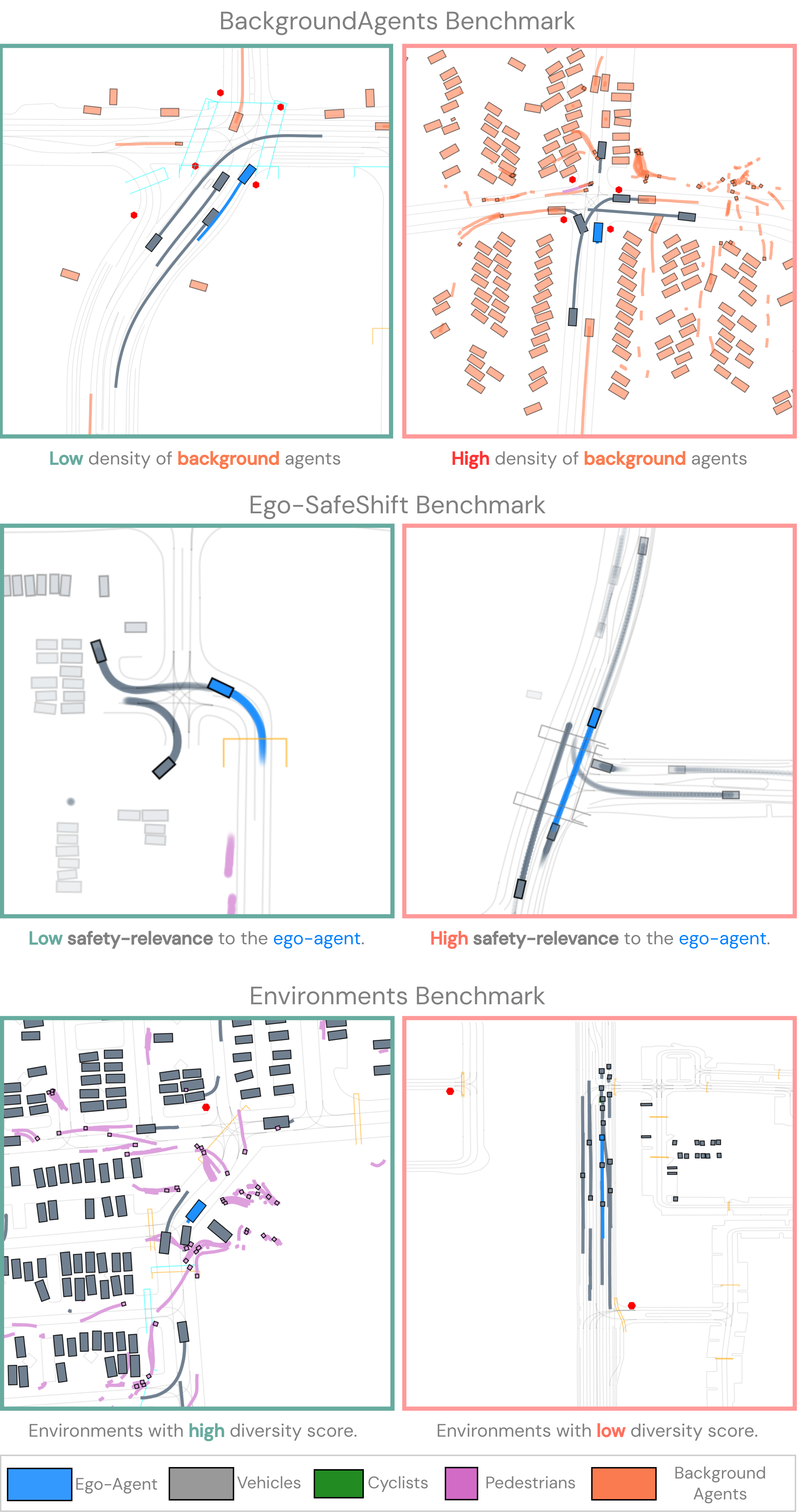}
    \caption{Representative \seen~(left, green panes) and \unseen~(right, red panes) scenarios for each benchmark: \backgroundagents~(top), \egosafeshift~(center), and \environments~(bottom). \seen~scenarios are used for training and validation; \unseen~scenarios are held out for testing.
    }
    \label{fig:distribution_shift_benchmarks}
\end{figure}

\subsection{\backgroundagents~Benchmark}
\label{ssec:background_agents}

This benchmark builds on \causalagents~\cite{sun2024causalagents}, in which human annotators labeled the agents that are causal to the \egoagent's behavior, leaving the remaining ones as \textit{background}~agents. These annotations enable scenario perturbations to mask out background agents, an intervention that should ideally leave predictions for the \egoagent~unchanged. Whereas Sun \etal~\cite{sun2024causalagents} apply this perturbation uniformly to measure degradation, we couple it with our formulation.
Specifically, the characterization function $c_{\backgroundagentsabv}$ counts the number of background agents in scenario $s$, serving as a proxy for perturbation magnitude: the more background agents a scenario contains, the larger the change induced by masking them. The splitting function $f_\backgroundagentsabv$ then holds out the top 15\% highest-count scenarios, whose perturbed counterparts form $\unseenS$, while the remaining scenarios constitute $\seenS$ in their original form.

The top row of \Cref{fig:distribution_shift_benchmarks} shows two scenarios with background agents labeled in orange: the left sub-figure (\seen) contains few, while the right one (\unseen) contains many.

\subsection{\egosafeshift~Benchmark}
\label{ssec:ego_safeshift}

This benchmark adapts the scenario characterization scheme of \safeshift~\cite{stoler2024safeshift, navarro2026scenchar}. Here, $c_\egosafeshiftabv$ assigns each scenario a scalar safety-relevance score, and $f_\egosafeshiftabv$ thresholds this score at the 85th percentile: the top 15\% highest-scoring scenarios form $\unseenS$, while the remainder constitutes $\seenS$ and is further partitioned into training and validation sets. While the original \safeshift~formulation focuses on capturing global safety-related properties, our objective is to capture finer-grained characteristics that increase decision-making difficulty for the \egoagent, so we modify the characterization to re-score existing scenarios from the \egoagent's perspective.

The center row of \Cref{fig:distribution_shift_benchmarks} shows two representative scenarios. In the left sub-figure (\seen), nearby agents pose low criticality to the \egoagent. In the right sub-figure (\unseen), the \egoagent~must account for multiple agents whose maneuvers intersect its intended path, increasing safety relevance and decision complexity. In both, higher agent opacity reflects higher relevance to the \egoagent.

\subsection{\environments~Benchmark}
\label{ssec:environments}

This benchmark induces a topological shift by holding out scenarios with structurally distinct maps. 
Here, we design a characterization function $c_\environmentsabv$ that converts each scenario map into a \networkx~\cite{hagberg2008networkx} graph and embeds it with \netlsd~\cite{tsitsulin2018netlsd}, a heat-trace graph descriptor satisfying four desirable properties: permutation invariance, scale adaptivity (capturing both local and global structure), size invariance, and scalability. 
The splitting function $f_\environmentsabv$ then applies agglomerative clustering over the resulting embeddings, using the silhouette score as a proxy for diversity and clusterability. The clusters with the highest silhouette scores are assigned to $\unseenS$: a cleanly separated cluster is internally homogeneous and structurally distinct from the rest of the corpus, and is therefore precisely its low-diversity, topologically atypical region. Training thus draws on the diverse remainder, and testing asks whether the structure learned there transfers to these narrower topologies.

The bottom row of \Cref{fig:distribution_shift_benchmarks} shows two representative scenarios: the left sub-figure (\seen) depicts a complex urban scenario with multiple crossing roads and parking regions, while the right sub-figure (\unseen) shows a high-speed straight corridor with few crossing points.

\section{Unified Robustness Evaluation}
\label{sec:evaluation}

\begin{table*}[t]
\centering
\small
\setlength{\tabcolsep}{4pt}
\vspace{0.2cm}
\caption{Model performance under each controlled distribution shift, one block per benchmark. We show the \seen~(validation) split followed by the \unseen~(test) split, with the parenthesized gap relative to \seen~in percent (red = degradation, green = improvement). \uniform~is the no-shift setting. The rightmost columns show per-benchmark robustness scores; a score of $1$ represents the reference. By design: \naive~scores $1.000$ on quality, being the quality reference, and all models score $1.000$ on stability under \uniform, being referenced against their own score there. Bold marks the best model per group column. Displacement error metrics are in meters, and CollisionRate uses a 0.25\,m threshold. Shaded rows are per-benchmark means. }
\label{tab:distribution_shift_results}
\resizebox{\textwidth}{!}{%
\begin{tabular}{l l cccccccccccccccc}
\toprule
\multirow{2}{*}{\textbf{Benchmark}} & \multirow{2}{*}{\textbf{Model}} & \multicolumn{5}{c}{\textbf{In Distribution (\seen, $\downarrow$)}} & \multicolumn{7}{c}{\textbf{Distribution Shift (\unseen, $\downarrow$)}} & \multicolumn{4}{c}{\textbf{Robustness Scores ($\uparrow$)}} \\
& & BrierFDE & MinFDE$_6$ & MinADE$_6$ & MissRate & CollisionRate &  &  & BrierFDE & MinFDE$_6$ & MinADE$_6$ & MissRate & CollisionRate &  &  & Quality & Stability \\
\midrule
\multirow{5}{*}{\uniform} & \naive & 11.228 & 10.561 & 3.649 & 0.640 & 0.048 &  &  & 11.357 (\textcolor{OrangeRed!100}{+1.15\%}) & 10.692 (\textcolor{OrangeRed!99}{+1.24\%}) & 3.689 (\textcolor{OrangeRed!94}{+1.09\%}) & 0.632 (\textcolor{ForestGreen!20}{-1.28\%}) & 0.046 (\textcolor{ForestGreen!41}{-4.18\%}) &  &  & 1.000 & 1.000 \\
 & \autobot & 3.213 & 2.519 & 2.480 & \textbf{0.368} & \textbf{0.017} &  &  & 3.195 (\textcolor{ForestGreen!20}{-0.58\%}) & 2.500 (\textcolor{ForestGreen!20}{-0.74\%}) & 2.453 (\textcolor{ForestGreen!20}{-1.05\%}) & \textbf{0.375} (\textcolor{OrangeRed!59}{+1.85\%}) & \textbf{0.016} (\textcolor{ForestGreen!20}{-7.27\%}) &  &  & 2.758 & 1.000 \\
 & \scenetransformer & 4.748 & 4.276 & 1.547 & 0.491 & 0.067 &  &  & 4.793 (\textcolor{OrangeRed!90}{+0.94\%}) & 4.330 (\textcolor{OrangeRed!100}{+1.26\%}) & 1.566 (\textcolor{OrangeRed!100}{+1.24\%}) & 0.490 (\textcolor{ForestGreen!32}{-0.31\%}) & 0.069 (\textcolor{OrangeRed!100}{+3.96\%}) &  &  & 1.837 & 1.000 \\
 & \wayformer & \textbf{2.792} & \textbf{2.306} & \textbf{0.880} & 0.385 & 0.060 &  &  & \textbf{2.806} (\textcolor{OrangeRed!70}{+0.51\%}) & \textbf{2.327} (\textcolor{OrangeRed!86}{+0.91\%}) & \textbf{0.887} (\textcolor{OrangeRed!83}{+0.77\%}) & 0.404 (\textcolor{OrangeRed!100}{+4.99\%}) & 0.061 (\textcolor{OrangeRed!94}{+3.21\%}) &  &  & \textbf{3.034} & 1.000 \\
 & \mtr & 3.839 & 3.316 & 1.210 & 0.424 & 0.054 &  &  & 3.841 (\textcolor{OrangeRed!48}{+0.04\%}) & 3.323 (\textcolor{OrangeRed!57}{+0.20\%}) & 1.209 (\textcolor{ForestGreen!53}{-0.09\%}) & 0.428 (\textcolor{OrangeRed!49}{+1.00\%}) & 0.056 (\textcolor{OrangeRed!95}{+3.31\%}) &  &  & 2.306 & 1.000 \\
\rowcolor[gray]{0.95}
 & Mean & 5.164 & 4.595 & 1.953 & 0.462 & 0.049 &  &  & 5.198 (+0.41\%) & 4.634 (+0.57\%) & 1.961 (+0.39\%) & 0.466 (+1.25\%) & 0.050 (-0.19\%) &  &  & 2.187 & 1.000 \\
\midrule
\multirow{5}{*}{\backgroundagents} & \naive & 12.019 & 11.364 & 4.819 & 0.668 & 0.037 &  &  & 13.395 (\textcolor{OrangeRed!20}{+11.44\%}) & 12.758 (\textcolor{OrangeRed!20}{+12.26\%}) & 5.129 (\textcolor{OrangeRed!20}{+6.45\%}) & 0.747 (\textcolor{OrangeRed!20}{+11.82\%}) & 0.033 (\textcolor{ForestGreen!61}{-12.46\%}) &  &  & 1.000 & 0.954 \\
 & \autobot & 3.319 & 2.625 & 2.474 & \textbf{0.380} & \textbf{0.023} &  &  & 4.038 (\textcolor{OrangeRed!88}{+21.67\%}) & 3.344 (\textcolor{OrangeRed!100}{+27.40\%}) & 3.222 (\textcolor{OrangeRed!86}{+30.22\%}) & 0.510 (\textcolor{OrangeRed!100}{+34.35\%}) & \textbf{0.026} (\textcolor{OrangeRed!100}{+12.38\%}) &  &  & 2.463 & 0.824 \\
 & \scenetransformer & 4.880 & 4.427 & 1.604 & 0.505 & 0.074 &  &  & 6.022 (\textcolor{OrangeRed!100}{+23.42\%}) & 5.452 (\textcolor{OrangeRed!77}{+23.15\%}) & 2.167 (\textcolor{OrangeRed!100}{+35.10\%}) & 0.659 (\textcolor{OrangeRed!86}{+30.52\%}) & 0.049 (\textcolor{ForestGreen!28}{-33.69\%}) &  &  & 1.852 & 0.915 \\
 & \wayformer & \textbf{2.901} & \textbf{2.426} & \textbf{0.922} & 0.414 & 0.067 &  &  & \textbf{3.486} (\textcolor{OrangeRed!78}{+20.19\%}) & \textbf{2.944} (\textcolor{OrangeRed!67}{+21.34\%}) & \textbf{1.131} (\textcolor{OrangeRed!65}{+22.67\%}) & 0.535 (\textcolor{OrangeRed!81}{+29.17\%}) & 0.042 (\textcolor{ForestGreen!23}{-36.79\%}) &  &  & \textbf{3.103} & 0.917 \\
 & \mtr & 3.753 & 3.237 & 1.188 & 0.399 & 0.057 &  &  & 4.550 (\textcolor{OrangeRed!85}{+21.24\%}) & 3.925 (\textcolor{OrangeRed!67}{+21.27\%}) & 1.477 (\textcolor{OrangeRed!69}{+24.33\%}) & \textbf{0.501} (\textcolor{OrangeRed!69}{+25.76\%}) & 0.035 (\textcolor{ForestGreen!20}{-39.13\%}) &  &  & 2.512 & \textbf{0.991} \\
\rowcolor[gray]{0.95}
 & Mean & 5.374 & 4.816 & 2.201 & 0.473 & 0.052 &  &  & 6.298 (+19.59\%) & 5.685 (+21.08\%) & 2.625 (+23.76\%) & 0.591 (+26.32\%) & 0.037 (-21.94\%) &  &  & 2.186 & 0.920 \\
\midrule
\multirow{5}{*}{\egosafeshift} & \naive & 11.480 & 10.961 & 3.672 & 0.643 & 0.049 &  &  & 11.810 (\textcolor{OrangeRed!20}{+2.88\%}) & 11.294 (\textcolor{OrangeRed!20}{+3.04\%}) & 3.965 (\textcolor{OrangeRed!20}{+7.99\%}) & 0.687 (\textcolor{OrangeRed!100}{+6.84\%}) & 0.053 (\textcolor{OrangeRed!20}{+8.04\%}) &  &  & 1.000 & 0.954 \\
 & \autobot & 3.456 & 2.762 & 2.912 & 0.422 & \textbf{0.022} &  &  & 4.043 (\textcolor{OrangeRed!100}{+16.96\%}) & 3.348 (\textcolor{OrangeRed!100}{+21.22\%}) & 3.584 (\textcolor{OrangeRed!100}{+23.08\%}) & 0.441 (\textcolor{OrangeRed!80}{+4.44\%}) & \textbf{0.032} (\textcolor{OrangeRed!100}{+44.71\%}) &  &  & 2.285 & 0.788 \\
 & \scenetransformer & 4.298 & 3.832 & 1.439 & 0.512 & 0.064 &  &  & 4.632 (\textcolor{OrangeRed!47}{+7.79\%}) & 4.142 (\textcolor{OrangeRed!42}{+8.09\%}) & 1.641 (\textcolor{OrangeRed!52}{+14.04\%}) & 0.514 (\textcolor{OrangeRed!47}{+0.41\%}) & 0.078 (\textcolor{OrangeRed!47}{+20.67\%}) &  &  & 1.980 & 1.016 \\
 & \wayformer & \textbf{2.592} & \textbf{2.113} & \textbf{0.828} & \textbf{0.374} & 0.052 &  &  & \textbf{2.847} (\textcolor{OrangeRed!59}{+9.83\%}) & \textbf{2.356} (\textcolor{OrangeRed!57}{+11.49\%}) & \textbf{0.904} (\textcolor{OrangeRed!26}{+9.24\%}) & \textbf{0.378} (\textcolor{OrangeRed!53}{+1.13\%}) & 0.066 (\textcolor{OrangeRed!61}{+26.93\%}) &  &  & \textbf{3.263} & \textbf{1.033} \\
 & \mtr & 3.704 & 3.185 & 1.156 & 0.433 & 0.055 &  &  & 4.032 (\textcolor{OrangeRed!53}{+8.84\%}) & 3.522 (\textcolor{OrangeRed!53}{+10.61\%}) & 1.312 (\textcolor{OrangeRed!48}{+13.46\%}) & 0.420 (\textcolor{ForestGreen!20}{-2.93\%}) & 0.070 (\textcolor{OrangeRed!62}{+27.51\%}) &  &  & 2.361 & 0.969 \\
\rowcolor[gray]{0.95}
 & Mean & 5.106 & 4.571 & 2.001 & 0.477 & 0.049 &  &  & 5.473 (+9.26\%) & 4.933 (+10.89\%) & 2.281 (+13.56\%) & 0.488 (+1.98\%) & 0.060 (+25.57\%) &  &  & 2.178 & 0.952 \\
\midrule
\multirow{5}{*}{\environments} & \naive & 8.409 & 8.042 & 2.699 & 0.677 & 0.050 &  &  & 8.054 (\textcolor{ForestGreen!20}{-4.22\%}) & 7.676 (\textcolor{ForestGreen!20}{-4.55\%}) & 2.709 (\textcolor{OrangeRed!20}{+0.38\%}) & 0.723 (\textcolor{OrangeRed!20}{+6.72\%}) & 0.046 (\textcolor{ForestGreen!38}{-7.30\%}) &  &  & 1.000 & \textbf{1.194} \\
 & \autobot & 3.224 & \textbf{2.529} & 2.359 & \textbf{0.371} & \textbf{0.018} &  &  & 4.149 (\textcolor{OrangeRed!100}{+28.70\%}) & 3.455 (\textcolor{OrangeRed!100}{+36.58\%}) & 2.783 (\textcolor{OrangeRed!75}{+17.96\%}) & \textbf{0.410} (\textcolor{OrangeRed!100}{+10.68\%}) & \textbf{0.017} (\textcolor{ForestGreen!82}{-2.12\%}) &  &  & \textbf{2.098} & 0.918 \\
 & \scenetransformer & 4.335 & 3.877 & 1.441 & 0.512 & 0.073 &  &  & 5.059 (\textcolor{OrangeRed!70}{+16.69\%}) & 4.555 (\textcolor{OrangeRed!62}{+17.49\%}) & 1.808 (\textcolor{OrangeRed!100}{+25.51\%}) & 0.553 (\textcolor{OrangeRed!46}{+8.04\%}) & 0.066 (\textcolor{ForestGreen!20}{-9.46\%}) &  &  & 1.462 & 0.982 \\
 & \wayformer & \textbf{3.200} & 2.717 & \textbf{1.040} & 0.455 & 0.053 &  &  & \textbf{3.452} (\textcolor{OrangeRed!49}{+7.88\%}) & \textbf{2.934} (\textcolor{OrangeRed!44}{+7.99\%}) & \textbf{1.145} (\textcolor{OrangeRed!51}{+10.18\%}) & 0.498 (\textcolor{OrangeRed!76}{+9.50\%}) & 0.053 (\textcolor{ForestGreen!100}{-0.09\%}) &  &  & 2.022 & 0.888 \\
 & \mtr & 3.769 & 3.240 & 1.205 & 0.407 & 0.052 &  &  & 4.045 (\textcolor{OrangeRed!48}{+7.32\%}) & 3.486 (\textcolor{OrangeRed!43}{+7.59\%}) & 1.321 (\textcolor{OrangeRed!49}{+9.64\%}) & 0.439 (\textcolor{OrangeRed!40}{+7.72\%}) & 0.048 (\textcolor{ForestGreen!33}{-7.90\%}) &  &  & 1.843 & 1.009 \\
\rowcolor[gray]{0.95}
 & Mean & 4.587 & 4.081 & 1.749 & 0.484 & 0.049 &  &  & 4.952 (+11.28\%) & 4.421 (+13.02\%) & 1.953 (+12.73\%) & 0.525 (+8.53\%) & 0.046 (-5.38\%) &  &  & 1.685 & 0.998 \\
\bottomrule
\end{tabular}%
}
\vspace{-0.4cm}
\end{table*}

\subsection{Experimental Setup}
\label{ssec:experimental_setup}

\paragraphbf{Dataset}
We validate our approach on the Waymo Open Motion Dataset (\womd)~\cite{ettinger2021large}. \womd~is among the largest and most diverse driving datasets available, spanning a wide range of roadway conditions and scene complexity. This makes it well suited for controlled shift induction along several axes. We use a subset of $\sim$45K scenarios. In each distribution shift setting of \Cref{sec:benchmark_suite}, we re-split this subset into \seen~for training and validation and \unseen~for testing, with a $70\%\ /\ 15\%\ /\ 15\%$ split, respectively. As a baseline for the introduced benchmarks, we include an experimental setting called \uniform, which applies no characterization function and instead splits the data uniformly across subsets. Every model is trained once per benchmark, so all reported differences are attributable to the induced shift.

\paragraphbf{Baselines}
We experiment on four widely used transformer-based baselines spanning diverse capacities: \autobot~\cite{girgis2021autobot} (1.5M parameters), \scenetransformer~\cite{ngiam2021scene} (7.6M), \wayformer~\cite{nayakanti2023wayformer} (15.1M), and \mtr~\cite{shi2022motion} (27.2M\footnote{Due to computational constraints, the \mtr~model used in our experiments is reduced from its original 60M parameters.}). Our reference model, \naive~(624k parameters), is a two-layer encoder-decoder predictor trained on \egoagent~trajectory data alone, \idest without map or social context.

\paragraphbf{Metrics}
We adopt standard trajectory forecasting metrics~\cite{feng2024unitraj}: Minimum Average and Final Displacement Error (MinADE$_K$, MinFDE$_K$) report, over $K=6$ predicted modes, the smallest distance to the ground-truth future averaged across all predicted timesteps and at the final timestep, respectively; BrierFDE extends MinFDE with a confidence penalty of $(1-p)^2$; MissRate is the fraction of scenarios in which no predicted trajectory falls within a distance threshold of the ground truth; and CollisionRate the fraction in which a predicted trajectory comes within a distance threshold of another agent's ground-truth trajectory.

\subsection{Per-Benchmark Results}
\label{ssec:model_results}

\Cref{tab:distribution_shift_results} contains one block per benchmark from \Cref{sec:benchmark_suite}, plus a block for the \uniform~setting of \Cref{ssec:experimental_setup}. Each block first reports performance on the validation (\seen) set, whose scenarios resemble those encountered during training, and then on the test (\unseen) set, along with the gap between the two as a percentage of in-distribution performance. The rightmost columns give the per-benchmark robustness scores defined in \Cref{ssec:robustness_scores}.

Models under \uniform~condition exhibit minimal performance loss, averaging less than $1\%$, while the proposed shifts induce significant degradation. Under \backgroundagents, predictions rarely collide with non-background agents, reflected in a marked drop in collision rate, yet they exhibit substantial behavioral degradation, as indicated by the rise in displacement error. Similarly, the \environments~benchmark shows an overall reduction in collision rate but an increase in displacement error, suggesting that while the held-out environments are less diverse in the interaction understanding and maneuvering they demand (as in \Cref{fig:distribution_shift_benchmarks}), they still require sound understanding of longitudinal operation, \exempli braking and accelerating. \egosafeshift shows the smallest gap in the FDE-based metrics but the largest increase in collision rate. This confirms the intuition of Stoler \etal~\cite{stoler2024safeshift}: the \unseen~set surfaces latent criticality, demanding a model's grasp of safety relevance and nuanced operation.


\subsection{Robustness Score Results}
\label{ssec:score_results}

\begin{figure}[!ht]
    \centering
    \vspace{0.2cm}
    \includegraphics[width=\linewidth]{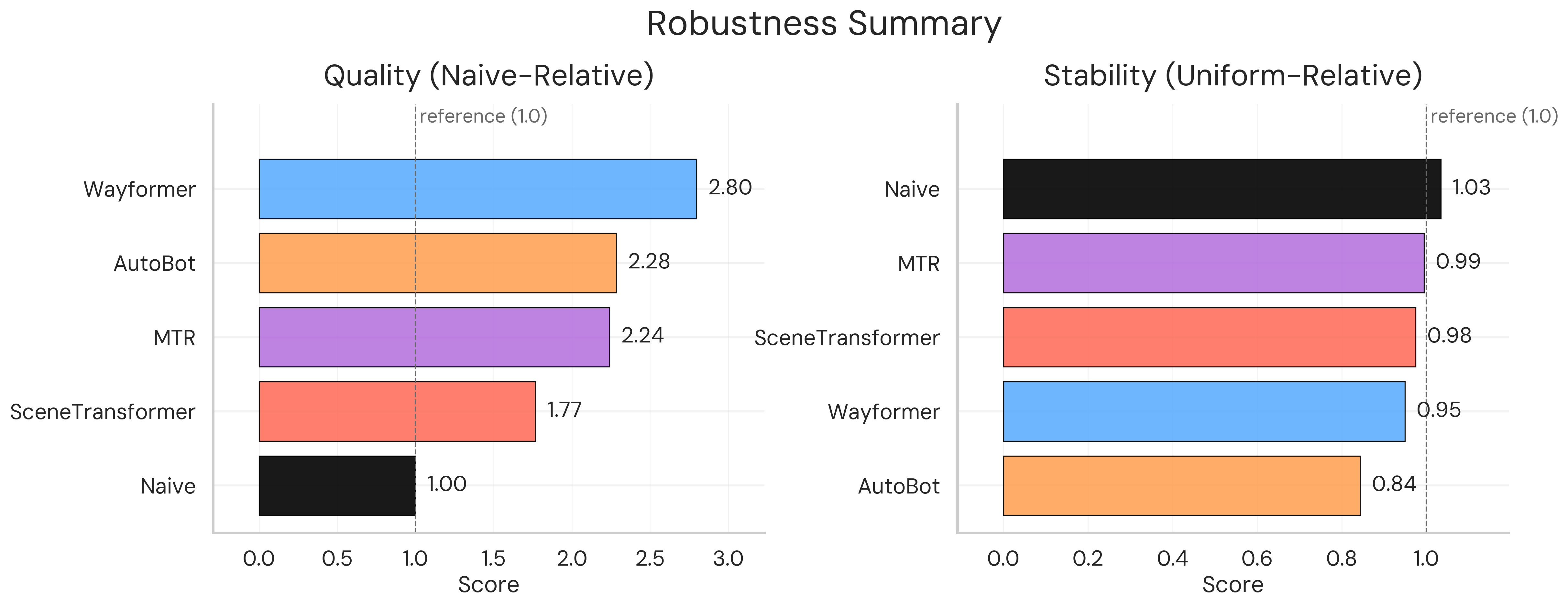}
    \caption{Robustness Scores, showing quality (left) and stability (right) rankings across models from \Cref{tab:distribution_shift_results}. Higher is better.}
    \label{fig:model_scores}
    \vspace{-0.5cm}
\end{figure}

\Cref{fig:model_scores} reports the unified robustness score of each model along the two dimensions of \Cref{ssec:unified_robustness}: \textit{quality} (left) and \textit{stability} (right), each aggregated over the three shifted benchmarks. The two rankings are close to inverses of one another. While this is not by construction, \idest a predictor could lead on both dimensions, this validates the change of reference captures complementary aspects of robustness rather than a single underlying ordering.

\paragraphbf{Quality}
\wayformer~attains the highest quality score, outperforming on displacement error metrics across benchmarks. However, like \scenetransformer~and \mtr, it falls below the \naive~reference model on collision rate. This suggests that the accuracy gains of high-capacity models are largely confined to the geometry of the modal trajectory and do not translate into proportionate improvements in safety-aligned metrics. \autobot, nonetheless, outperforms all in collision rate, albeit exhibiting a larger relative degradation under distribution shift (\Cref{tab:distribution_shift_results}), indicating that relative degradation alone can be a misleading proxy for downstream reliability.

\paragraphbf{Stability}
Here each model is referenced against \textit{its own} performance in the \uniform~setting. Although the learned models are all substantially stronger than \naive~in absolute downstream performance, each degrades more than \naive~under distribution shift. Except for collision rate, degradation is broadly uniform across metrics and models under \seen~conditions. Under \unseen~conditions, the pattern inverts relative to the quality axis: \wayformer, \mtr, and \scenetransformer~lose displacement error stability against the reference while improving on collision rate. \autobot~alone degrades on every metric, obtaining the lowest combined stability score.

\section{Discussion}
\label{sec:discussion}

\subsection{Conclusion}
\label{ssec:conclusion}

We presented \framework, a framework and benchmark suite for standardized evaluation of trajectory prediction robustness under distribution shift, together with a unified robustness score that condenses performance across benchmarks into interpretable quality and stability values. 
Together, these components provide a consistent and principled protocol for directly comparing methods. 

We hope our framework helps the community contextualize robustness across methods and shift conditions, guiding the development of models that respond sensibly to novel situations that autonomous systems will inevitably encounter at deployment.

\subsection{Limitations and Future Work}
\label{ssec:future_work}

We see three main avenues for future work, each addressing current limitations. First, our robustness score is defined relative to an anchor~\cite{hyndman2006another, koh2021wilds, sagawa2020distributionally}, which is non-trivial to choose. A constant-velocity model, for instance, is principled, but its limited performance reduces the range of meaningful comparison, making robustness gaps less informative. A naive learning-based model yields stronger performance and a more interesting basis for comparison, at the cost of interpretability. Establishing anchors that retain principled design without sacrificing interpretability remains an open problem.

Second, our three axes of variation, chosen for their relevance to the \egoagent, do not exhaust the space of shifts an autonomous system may encounter. Because our formulation only requires specifying $c$ and $f$, additional axes such as geographic transfer~\cite{caesar2020nuscenes} or agent-type composition~\cite{stoler2025longcomp, kothari2023motion} could be instantiated within the same protocol.

Third, we instantiated \framework~only in the domain of driving. Generalizing it to other large-scale domains such as aviation~\cite{patrikar2022trajair, navarro2022social, navarro2024amelia} is therefore a promising direction, which would additionally admit axes of analysis such as time-of-day and weather, which are not generally available in trajectory-based driving datasets.


\addtolength{\textheight}{0cm}
\appendices

\section{Benchmarks Analysis}
\label{sec:benchmark_analysis}

This section provides data analyses that further validate the proposed benchmarks.

\subsection{\backgroundagents~Analysis} 

\begin{figure}[!ht]
    \centering
    \includegraphics[width=0.96\linewidth]{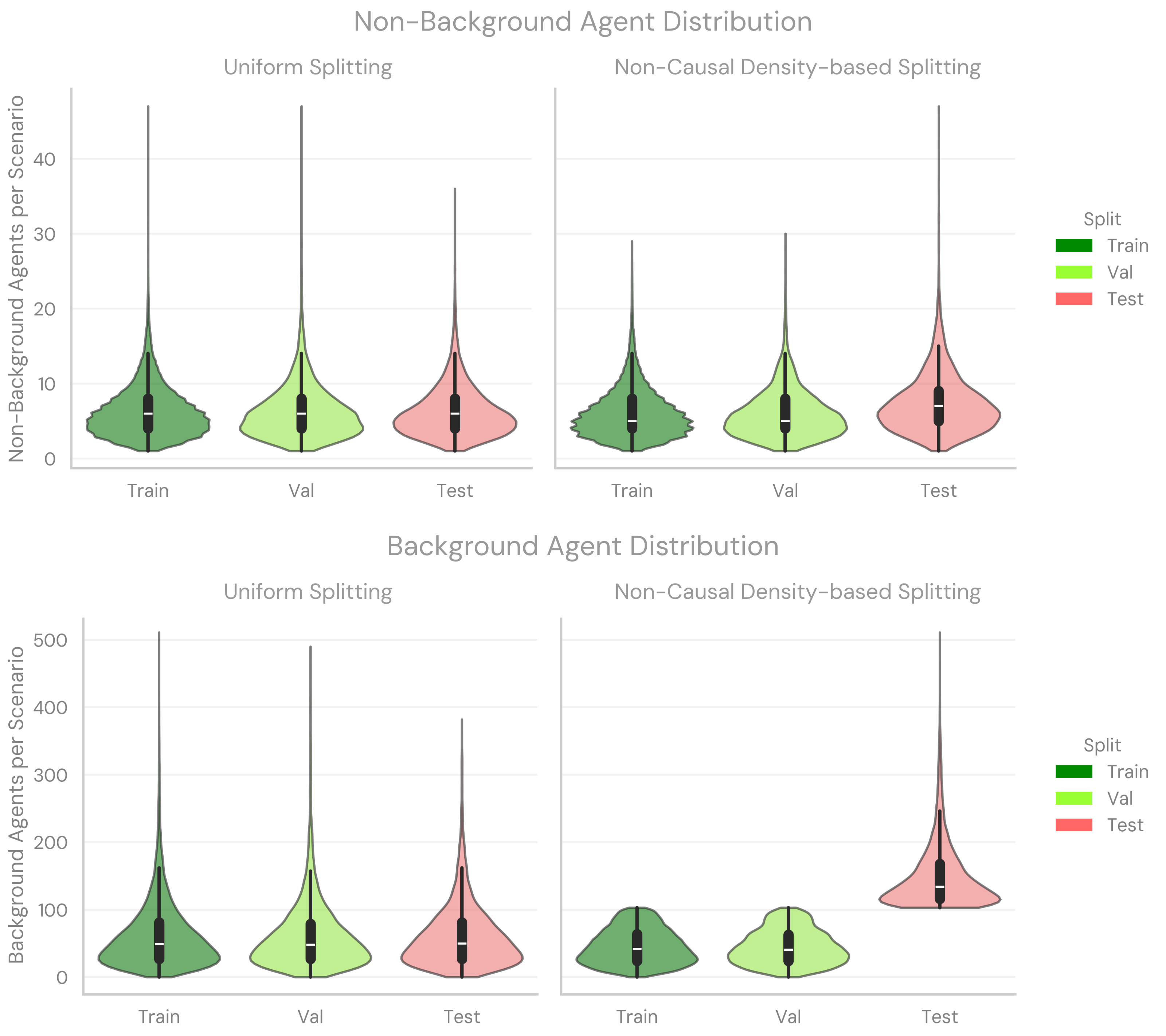}
    \caption{Distribution of \textit{non-background} (top) and \textit{background} (bottom) agents per scenario across splits, comparing uniform splitting (left) with our background, density-based splitting (right) for the \backgroundagents~benchmark. 
    While the non-background distribution is preserved across both strategies, our splitting concentrates high background agent density in the test set.
    }
    \label{fig:background_agents_analysis}
\end{figure}

\Cref{fig:background_agents_analysis} shows the distribution of \textit{background} and \textit{non-background} agents per scenario across data splits. Each row compares two splitting strategies: \textit{uniform} (left) and the \textit{background, density-based} splitting of \Cref{ssec:background_agents} (right).

The top row reports the distribution of non-background agents, which ranges roughly from 0 to 50 agents per scenario. In both strategies, the distributions remain similar across splits, confirming that our proposed splitting does not distort the distribution of relevant agents that models must reason over.

The bottom row reports the distribution of \textit{background} agents, which ranges roughly from 0 to 500 agents per scenario. Under uniform splitting, the distributions remain similar. Under the proposed one, however, the test set is deliberately shifted toward scenarios with substantially higher background agent density. This shift exposes the model at test time to scenarios densely populated with agents that do not directly influence its decision-making, providing a more demanding evaluation of robustness to non-influential agents.

\subsection{\egosafeshift~Analysis} 

\begin{figure}[!ht]
    \centering
    \includegraphics[width=0.98\linewidth]{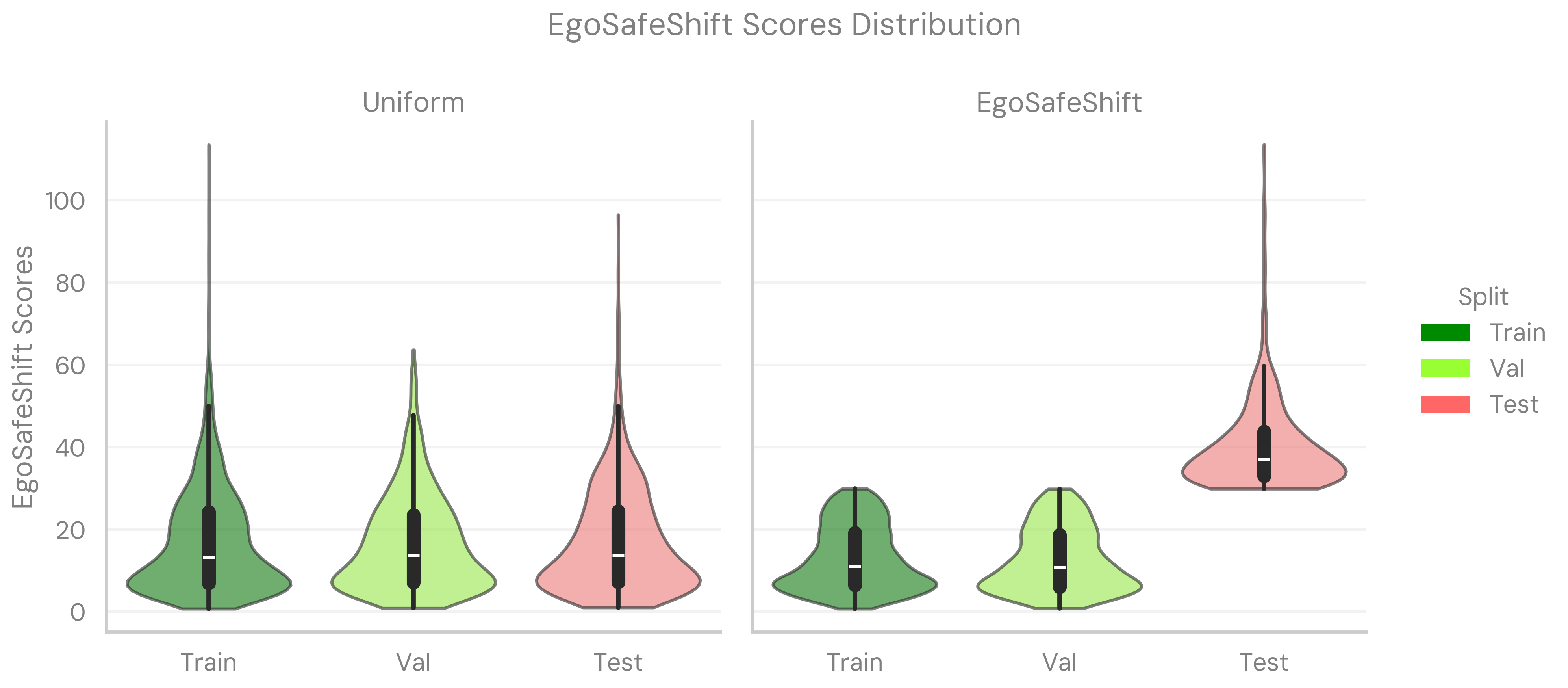}
    \caption{Distribution of \textit{safety-relevant scores} per scenario across splits, comparing uniform splitting (left) with our \egosafeshift~re-splitting (right).
    Under \uniform~splitting, all three splits have similar score distributions. Under our splitting the train/val distributions concentrate on low-scored scenarios, whereas the test distribution concentrates on higher-scored scenarios where latent criticality is more relevant.
    }
    \label{fig:ego_safeshift_analysis}
\end{figure}

\Cref{fig:ego_safeshift_analysis} shows the distribution of \textit{safety-relevant scores} per scenario across splits. Scores range from 0 to over 100, where a higher score indicates a scenario that is more challenging for the \egoagent, as defined by the \egosafeshift~benchmark proposed in \Cref{ssec:ego_safeshift}. The left subplot shows the distribution under \textit{uniform} splitting, while the right shows the distribution under our proposed re-splitting.

Under uniform splitting, the three splits are similarly distributed, centered around low-score regimes. Under our proposed splitting, the test set is shifted toward higher-scoring scenarios. By concentrating these more difficult scenarios in the test set, our splitting evaluates the model on nuanced, preemptive decision-making and latent criticality, providing a more demanding test of safety-relevant robustness.

\subsection{\environments~Analysis} 

\begin{figure}[!ht]
    \centering
    \vspace{0.2cm}
    \includegraphics[width=0.98\linewidth]{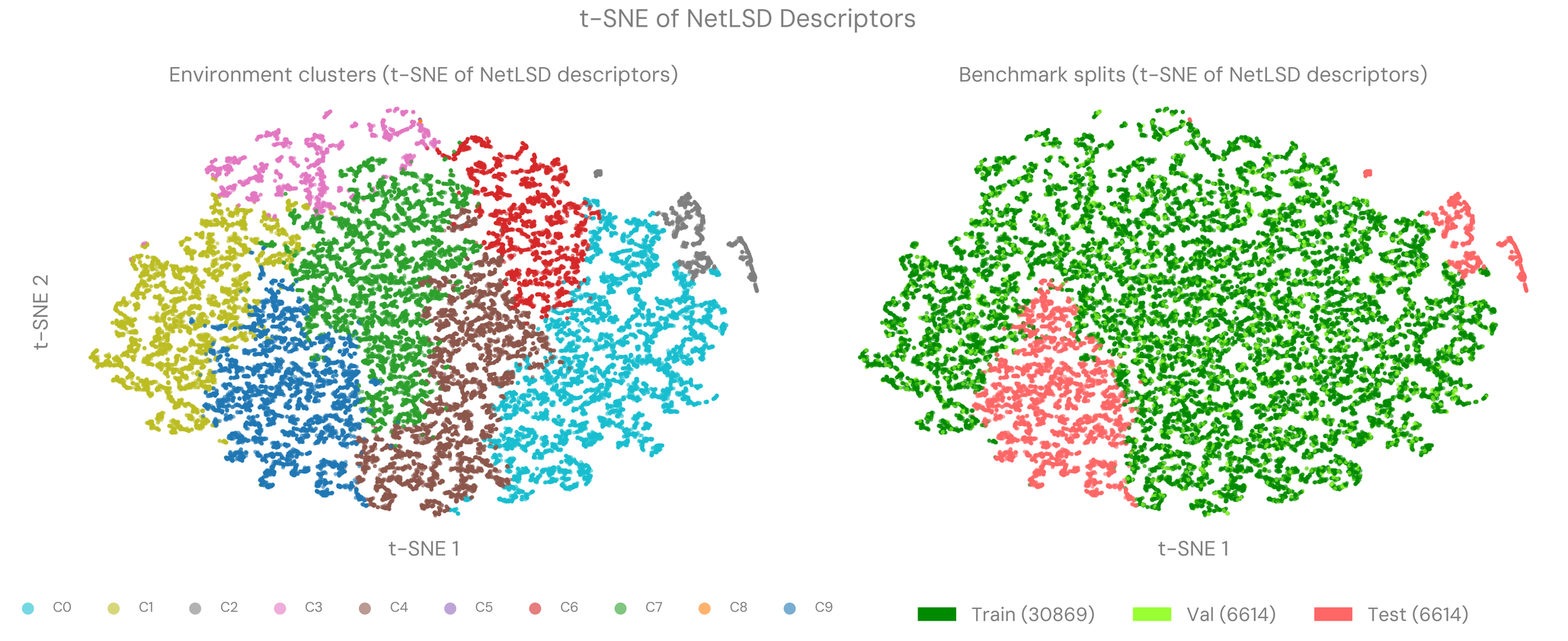}
    \caption{Distribution of \netlsd descriptors for the \environments~benchmark using a 2D t-SNE projection. The panes show the projection colored by cluster assignment (left) and by data split (right). 
    Under our splitting, the clusters with the highest silhouette score are assigned to the test set, as this score is used as a proxy for separability of map topologies. 
    }
    \label{fig:environments_analysis}
\end{figure}

\Cref{fig:environments_analysis} shows the scenario distribution derived from the environment embeddings produced with the \netlsd~\cite{tsitsulin2018netlsd} descriptor, as described in \Cref{ssec:environments}. To aid visualization, we plot a t-SNE projection of the embeddings. The left panel colors each embedding by its cluster assignment; in our experiments, we used 10 clusters. The right panel colors the embeddings by split assignment. We assign the clusters with the highest silhouette scores to the test set: a higher silhouette score indicates a cluster that is more cleanly separated from the rest of the corpus and more internally homogeneous, \idest lower in internal diversity, so we expect the environments in these clusters to have topologies that are both distinct from and narrower than those in the remaining clusters.

\begin{figure}[!t]
    \centering
    \includegraphics[width=0.98\linewidth]{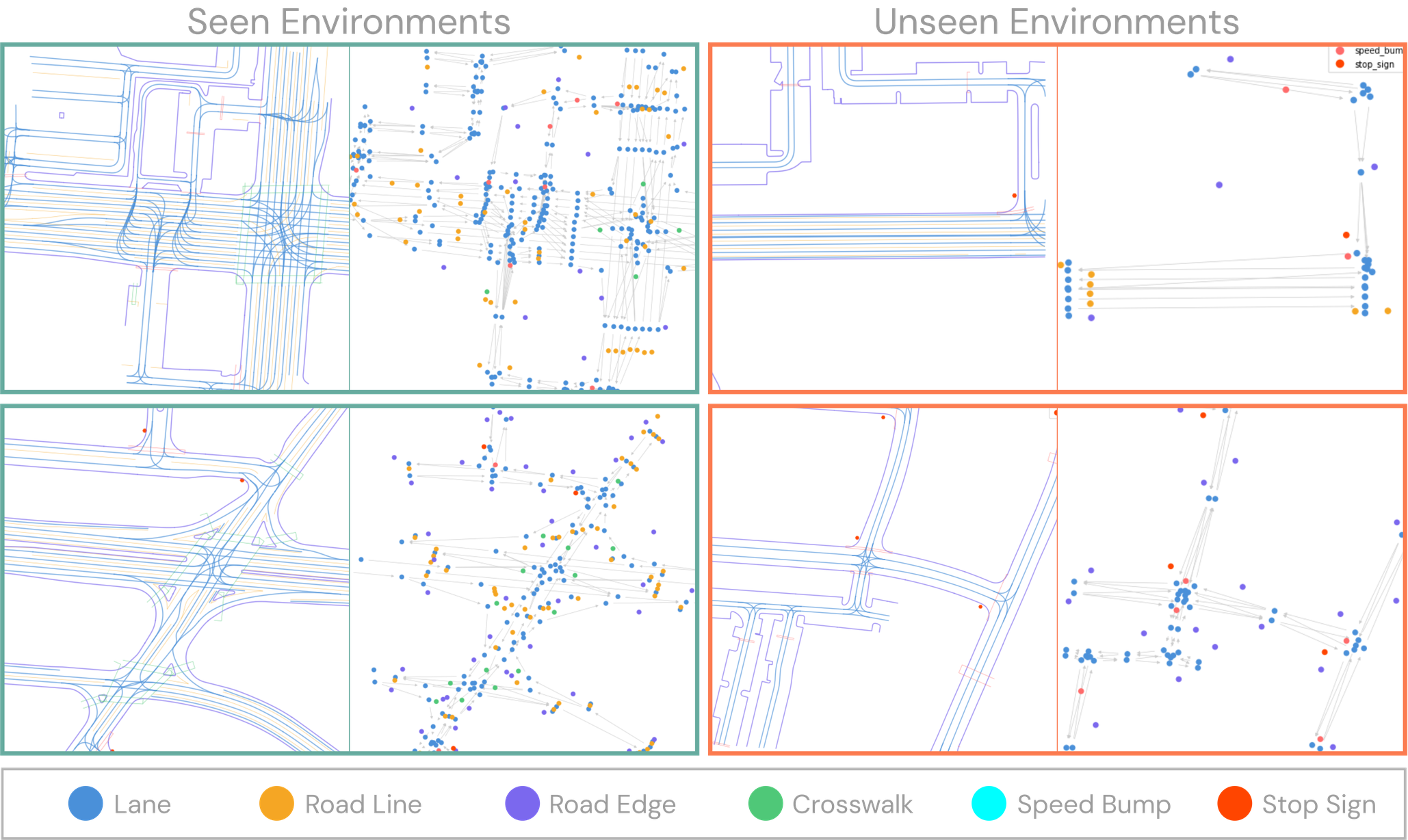}
    \caption{Examples of road topologies illustrating the shift induced by \environments. 
    Side-by-side, we show the road layout and \netlsd-encoded graph, both colored according to the legend. \seen environments (left) show dense, highly connected layouts with multi-lane intersections and wide avenues.
    \unseen~environments (right) show sparse, weakly connected layouts of long parallel corridors requiring primarily straight-line travel. This graph density difference is what the \netlsd heat trace captures and what our clustering separates on. 
    }
    \label{fig:graph_analysis}
\end{figure}

In \Cref{fig:graph_analysis}, we show four examples of road layouts and their corresponding graphs encoded with \netlsd to illustrate the structural differences across scenarios in the \seen~(top) and \unseen~(bottom) sets. The \seen~examples exhibit highly complex environments that require understanding multiple maneuver types, \exempli turning, moving straight, navigating intersections, and traversing wide avenues, all of which yield dense graphs. In contrast, the \unseen~examples show less complex environments with sparser graphs, requiring primarily straight-line navigation.

\subsection{Benchmark Overlap} 

\begin{figure}[!htp]
    \centering
    \vspace{0.2cm}
    \includegraphics[width=0.96\linewidth]{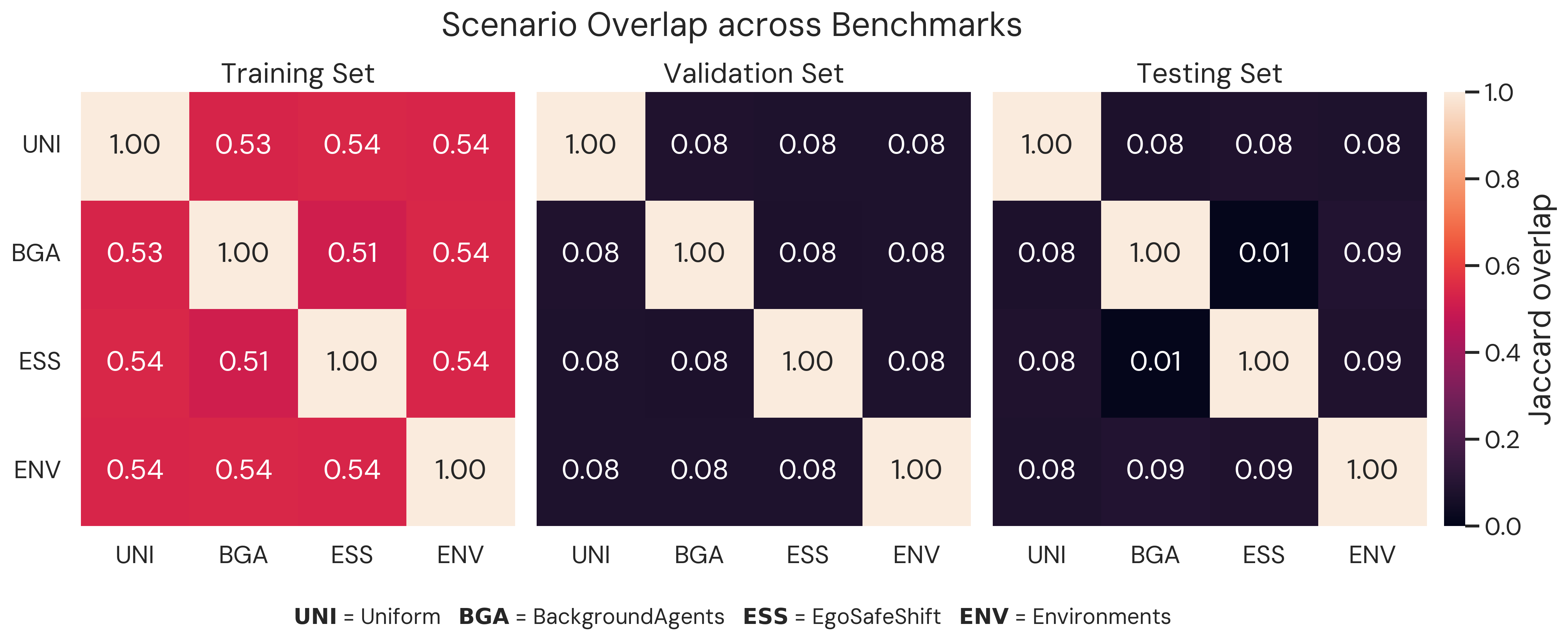}
    \caption{Pairwise Jaccard overlap of scenarios across benchmarks, shown separately for training, validation, and test splits. High training overlap reflects shared scenario pools, while consistently low validation and test overlap demonstrates that the benchmarks evaluate disentangled conditions.}
    \label{fig:benchmark_overlap}
    \vspace{-0.3cm}
\end{figure}

As a proxy for benchmark-to-benchmark disentanglement, we report a pairwise overlap analysis across benchmarks in \Cref{fig:benchmark_overlap}. The left heatmap shows overlap across training sets, where benchmark pairs share slightly more than 50\% of their data. This is expected as training sets are large and drawn from shared scenario pools; the disentanglement we care about is on the held-out sets. The center heatmap shows overlap across validation sets, which is uniformly low at 8\% across all pairs. The right heatmap shows the test overlap: \uniformabv~shares 8\% with every other benchmark, {\small{\backgroundagentsabv-\environmentsabv}}~and {\small{\egosafeshiftabv-\environmentsabv}}~share 9\%, and {\small{\backgroundagentsabv-\egosafeshiftabv}}~shares only 1\%. The consistently low validation and test overlap indicates that each benchmark evaluates a sufficiently unique and disentangled set of conditions.

\ifanonymize
\else
\section*{Acknowledgment}
This work was done in collaboration with Stack AV and Lavoro AI Research; we thank them for their mentorship. 
\fi


\bibliographystyle{IEEEtran}
\bibliography{ref}


\end{document}